\documentclass[runningheads]{llncs}
\usepackage{xcolor}
\usepackage{graphicx}
\usepackage{bm}
\newcommand{\vect}[1]{\boldsymbol{\mathbf{#1}}}
\usepackage{caption}
\usepackage{amsmath,amssymb}
\usepackage{tabularray}
\UseTblrLibrary{booktabs}
\usepackage{multirow}
\usepackage{marvosym}
\usepackage{color, colortbl}
\usepackage{float}

\usepackage{hyperref}

\begin{document}

\title{Evaluating Reliability in Medical DNNs: A Critical Analysis of Feature and Confidence-Based OOD Detection}
\titlerunning{Evaluating Reliability in Medical DNNs}
\author{Harry Anthony$^{\textrm{1,(\Letter)}}$, Konstantinos Kamnitsas$^{1,2,3}$} 
\authorrunning{H. Anthony et al.}
%
\institute{{$^1$Department of Engineering Science, University of Oxford, Oxford, UK}  {\\ $^2$Department of Computing, Imperial College London, London, UK \\ $^3$School of Computer Science, University of Birmingham, Birmingham, UK }\email{harry.anthony@eng.ox.ac.uk}
}%
\maketitle              
\begin{abstract}
Reliable use of deep neural networks (DNNs) for medical image analysis requires methods to identify inputs that differ significantly from the training data, called out-of-distribution (OOD), to prevent erroneous predictions. OOD detection methods can be categorised as either \textit{confidence-based} (using the model's output layer for OOD detection) or \textit{feature-based} (not using the output layer). We created two new OOD benchmarks by dividing the D7P (dermatology) and BreastMNIST (ultrasound) datasets into subsets which either contain or don't contain an artefact (rulers or annotations respectively). Models were trained with artefact-free images, and images with the artefacts were used as OOD test sets. For each OOD image, we created a counterfactual by manually removing the artefact via image processing, to assess the artefact's impact on the model's predictions. We show that OOD artefacts can boost a model’s softmax confidence in its predictions, due to correlations in training data among other factors. This contradicts the common assumption that OOD artefacts should lead to more uncertain outputs, an assumption on which most confidence-based methods rely. We use this to explain why feature-based methods (e.g. Mahalanobis score) typically have greater OOD detection performance than confidence-based methods (e.g. MCP). However, we also show that feature-based methods typically perform worse at distinguishing between inputs that lead to correct and incorrect predictions (for both OOD and ID data). Following from these insights, we argue that a combination of feature-based and confidence-based methods should be used within DNN pipelines to mitigate their respective weaknesses. These project's code and OOD benchmarks are available at: \url{https://github.com/HarryAnthony/Evaluating_OOD_detection}.

\keywords{Out-of-distribution  \and Uncertainty \and Distribution shift.}
\end{abstract}
\section{Introduction}\label{Sec:Intro}
Deep Neural Nets (DNNs) have emerged as powerful tools for analysing medical images, and have been found promising for various tasks such as classifying diseases \cite{topol_high-performance_2019}. However, when they encounter data that differ significantly from the training data, out-of-distribution (OOD), their generalisation is unpredictable \cite{hendrycks_baseline_2018}. This motivated research in OOD detection, to create methods to identify when a model prediction is unreliable - mitigating risk of downstream errors.

We can separate OOD detection methods into two categories: \textit{internal methods} (methods that use the parameters or outputs of a DNN which has been trained for a specific task e.g. classification) and \textit{external methods} (external to the DNN).
\textit{External methods} encompass many approaches, such as one-class classifier methods \cite{scholkopf_support_1999,ruff_deep_2018}, density-based methods \cite{kobyzev_normalizing_2021,ren_likelihood_2019} and reconstruction-based methods \cite{tan_detecting_2021,liang_modality_2023}. \textit{Internal methods} can be further separated into those that do not require a DNN to be retrained (\textit{post-hoc methods}) and those that require a specific type of training (\textit{ad-hoc methods}). \textit{Ad-hoc methods} cover a broad spectrum of techniques, from altering the network's architecture (e.g. Bayesian Neural Networks \cite{tishby_consistent_1989,barber_ensemble_nodate} and confidence enhancement methods \cite{devries_learning_2018,van_amersfoort_uncertainty_2020}) to changing how a network is trained (e.g. outlier exposure \cite{hendrycks_deep_2018,liu_energy-based_2021}). This paper focuses on \textit{post-hoc internal methods}, which have several benefits: they can be applied to pre-trained networks, they typically don't have restrictions on architecture design and are typically low computational cost. We can further separate \textit{post-hoc internal methods} into \textit{confidence-based methods}, which use a model's output layer for OOD detection, and \textit{feature-based methods}, which use other features for OOD detection (e.g. hidden layer data) \cite{roschewitz_distance_2023}. 

The primary goal of integrating OOD detection methods into a DNN pipeline for image classification is to identify more \textit{trustworthy} predictions, which are likely to be accurate and less susceptible to unpredictable diagnoses caused by the model's interactions with OOD features. Most OOD detection studies evaluate methods on their ability to separate ID and OOD inputs, using a metric like AUROC \cite{ruff_unifying_2021}. However, there's a growing body of research that have argued the traditional OOD framework does not effectively reveal which method is best at detecting errors \cite{guerin_out--distribution_2023,zhu_rethinking_2022,gal_sena_2023,jaeger_call_2022}. They propose evaluating methods based on their ability to specifically discard incorrect predictions, regardless of whether these predictions are from OOD inputs (known as failure detection). 
To this end, we analyse the strengths and weaknesses of both feature-based and confidence-based methods at OOD detection and failure detection. By identifying the respective weaknesses of these methods, we consider how they can be used to make DNN predictions more trustworthy. We make the following contributions:

\begin{itemize}
    \item Develop two OOD benchmarks by categorising all images from the D7P and BreastMNIST datasets into those with and without artefacts (rulers and annotations respectively). We manually create modified versions of each of the 478 images with artefacts by removing the artefacts using a patch from the same image, allowing for an analysis of their impact on the model's predictions. As a contribution, we made this data publicly available.
    \item We challenge assumptions on which confidence and feature-based detection methods rely: OOD artefacts should lead to uncertain (high entropy) model outputs, and the distance of an input to the training data in the model's latent space is a reliable predictor of diagnosis accuracy.
    \item We use these false assumptions to explain and demonstrate that \textit{feature-based methods} typically perform better at OOD detection than \textit{confidence-based methods}, whereas \textit{confidence-based methods} typically perform better at failure detection than \textit{feature-based methods}.
    \item To our knowledge, be the first paper to motivate and demonstrate the benefit of combining both confidence and feature-based OOD detection methods in a DNN pipeline to mitigate their respective weaknesses.
\end{itemize}

\section{Material and Methods}\label{Sec:Methods}

\textbf{Primer on OOD detection.}
Consider a model's input, $\mathbf{x} \in \mathcal{X}$ and a label $y \in \mathcal{Y} =\{ 1,...,K \}$ from a label-space with K classes. Data used to train a network, $f$, is from a sample $\mathcal{D}_{\text {train}} = \{ (x_n, y_n) \}_{n=1}^N \subset \mathcal{X} \times \mathcal{Y}$. 
This paper studies covariate-shifted OOD data $p_{\text{train}}(\mathbf{x}) \neq p_{\text{test}}(\mathbf{x})$, which has the same label-space as the training data, but a distribution shift in $\mathbf{x}$ (i.e. unseen artefact) - sometimes referred to as near OOD \cite{yang_full-spectrum_2022}. 
OOD detection can be viewed as a binary classification problem, where we get a confidence scoring function from our method $\mathcal{S}(\mathbf{x},f)$ for an input $\mathbf{x}$.  We label $\mathbf{x}$ as OOD when the scoring function $\mathcal{S}(\mathbf{x},f)$ is below a threshold $\lambda$, and ID if it is above. 
The primary metric for evaluating the performance of OOD detection is AUROC$_{\text{OOD}}$, which assesses the scoring function's ability to distinguish ID from OOD inputs \cite{hendrycks_baseline_2018}. As outlined in Sec. \ref{Sec:Intro}, our focus extends to evaluating the scoring function's ability for failure detection, which we quantify using AUROC$_{f}$ \cite{jaeger_call_2022}. 
For \text{AUROC}$_{\text{OOD}}$ the true label is ID and the false label is OOD, and for \text{AUROC}$_{f}$ the true label is a correct diagnosis and false label is an incorrect diagnosis \cite{jaeger_call_2022}. 

\textit{Confidence-based} OOD detection methods use the model's output layer for OOD detection. An example is Maximum Class Probability (MCP) \cite{hendrycks_baseline_2018}, which uses the maximum class softmax probability as the scoring function 
\begin{equation}\label{eq:MCP}
\mathcal{S}_{\text {MCP}}(\mathbf{x}, f) = \max_{y \in \mathcal{Y}} \  \text{softmax}[ f(\mathbf{x} | \mathcal{D}_{\text {train}} ) ].
\end{equation}
Other examples of confidence-based methods include Shannon Entropy (SE) \cite{hendrycks_scaling_2022}, Max Logit Score (MLS) \cite{hendrycks_scaling_2022}, Energy score \cite{liu_energy-based_2021}, MCP from Monte Carlo Dropout (MCDP-MCP) \cite{gal_dropout_2016}, predicted entropy from  Dropout (MCDP-PE) \cite{jaeger_call_2022}, Mutual Information from Dropout (MCDP-MI) \cite{jaeger_call_2022}, MCP from Deep Ensembles (DE-MCP) \cite{lakshminarayanan_simple_2017} and GradNorm \cite{huang_importance_2021}. Some methods increase the separation between ID and OOD data by either increasing the model's confidence in a diagnosis (ODIN \cite{liang_enhancing_2020}) or reducing it (ReAct \cite{sun_react_nodate} and DICE \cite{sun_dice_2022}) - these methods have hyperparameters that can be optimised on a validation OOD set.

In contrast, \textit{feature-based} OOD detection methods don't use the model's output layer. An example is Mahalanobis score, which uses a feature extractor $\mathcal{F}$ (typically a section of the DNN) to extract feature maps from a hidden layer $h(\mathbf{x}) \in \mathbb{R}^{J \times J \times M}$, where the maps have size $J \times J$ with $M$ channels. The feature map's means can be used to define a vector $\mathbf{z}(\mathbf{x}) \in \mathbb{R}^{M} = \frac{1}{J^2} \sum_J \sum_J \mathbf{h} (\mathbf{x})$. Firstly, the mean $\vect{\mu_y}$ and covariance matrix $\Sigma_y$ of each class in the training data $(\mathbf{x},y) \sim \mathcal{D}_{\text {train}}$ is calculated.

The Mahalanobis distance between the vector $\mathbf{z}$($\mathbf{x}$) of a test data point $\mathbf{x}$ and the training data of class $y$ can be calculated as a sum over M dimensions. The Mahalanobis score $d_{\mathcal{M}}$ is defined as the minimum Mahalanobis distance between the test data point and the class centroids of the training data, and it can be used as an OOD scoring function \cite{lee_simple_2018}.
\begin{equation}\label{Eq:Mahal score}
    d_{\mathcal{M}_y}(\mathbf{x}) = \sum_{i=1}^M ( \mathbf{z}(\mathbf{x}) - \vect{\mu_y}) \Sigma_y^{-1}  ( \mathbf{z}(\mathbf{x}) - \vect{\mu_y}), \  \mathcal{S}_{\text {Mahal.}}(\mathbf{x},f) = - \min_{y \in \mathcal{Y}}  d_{\mathcal{M}_y}(\mathbf{x}) 
\end{equation}
\noindent Previous works suggest the OOD detection performance of Mahalanobis score could be improved by combining distances from different layers, called Multi-branch Mahalanobis (MBM) \cite{anthony_use_2023}, or by measuring the distance relative to the distribution of all the training data, called Relative Mahalanobis Score (RMS) \cite{ren_simple_2021}. Another feature-based method uses the differences in GRAM matrices \cite{sastry_detecting_2020}.

\noindent \textbf{Explainable AI (XAI) methods}\label{sec:XAI_methods}
 generate saliency maps to weight the relevance of pixels of an image for a model's diagnosis, aiming to make the model's decision-making process more understandable - an example method is Layer Relevance Propagation (LRP) \cite{montavon_layer-wise_2019}.

\textbf{Datasets and Implementation.}\label{Sec:OOD_methods}
We manually annotated two datasets - BreastMNIST (ultrasound images) \cite{yang_medmnist_2023} and D7P (dermatology images) \cite{kawahara_7-point_2018} - into subsets of images that contain an artefact (rulers and annotations respectively) and images that do not.  Models were trained on 90\% of the images without the artefact, with 10\% used as held-out ID test cases (table \ref{tbl:data}). These tasks were selected for OOD analysis because the artefacts do not provide clinically useful information for diagnosing pathology, and are easy to localise and remove. The models used were ResNet18 and VGG16, where we trained 5 seeds.


\begin{table}[ht]
\centering
\caption{Summary of ID and OOD data used for OOD detection evaluation.}
\begin{tblr}[t]{colspec={Q[l] | Q[l] Q[l] Q[l] | Q[l] Q[l]}, row{2,4}={bg=gray!10}, column{1}={bg=white!10}}
\toprule
Dataset & Classes & \# ID img & Train:Test & OOD Artefact &  \# OOD img  \\
\midrule
Breast- & Normal &  126 & \SetCell{bg=white} 90:10 & \SetCell{bg=white} Annotations & 7 \\
MNIST &	Benign & 269 &  &  & 168 \\
\cite{yang_medmnist_2023} &	Malignant & 157 & \SetCell{bg=white}  &  \SetCell{bg=white}  & 53 \\
\midrule
D7P & Nevus & 832 & \SetCell{bg=white} 90:10 & \SetCell{bg=white} Grid ruler  & 148\\
\cite{kawahara_7-point_2018} & \SetCell{bg=gray!10} Not Nevus  & \SetCell{bg=gray!10} 571 &  &  & \SetCell{bg=gray!10} 102 \\
\bottomrule
\end{tblr}
\label{tbl:data}
\end{table}

For each of the 478 OOD images, we made pixel-wise segmentation masks for the artefact. We then manually replaced the pixels of the artefact with pixels in the same image (intra-image interpolation), using a Gaussian smoothing filter to ensure smooth boundaries. This was chosen over using a pre-trained generative model to remove the artefact because we can ensure we are not introducing a new unexpected OOD artefact into an image, and to prevent changing the true label of the image. This was done to approximate the image without the artefact, allowing us to evaluate the impact of the artefact on a model's diagnosis. The dataset annotations, segmentation masks and synthetic image datasets have all been made publicly available.

\begin{figure}
\centering
 \includegraphics[width=11cm]{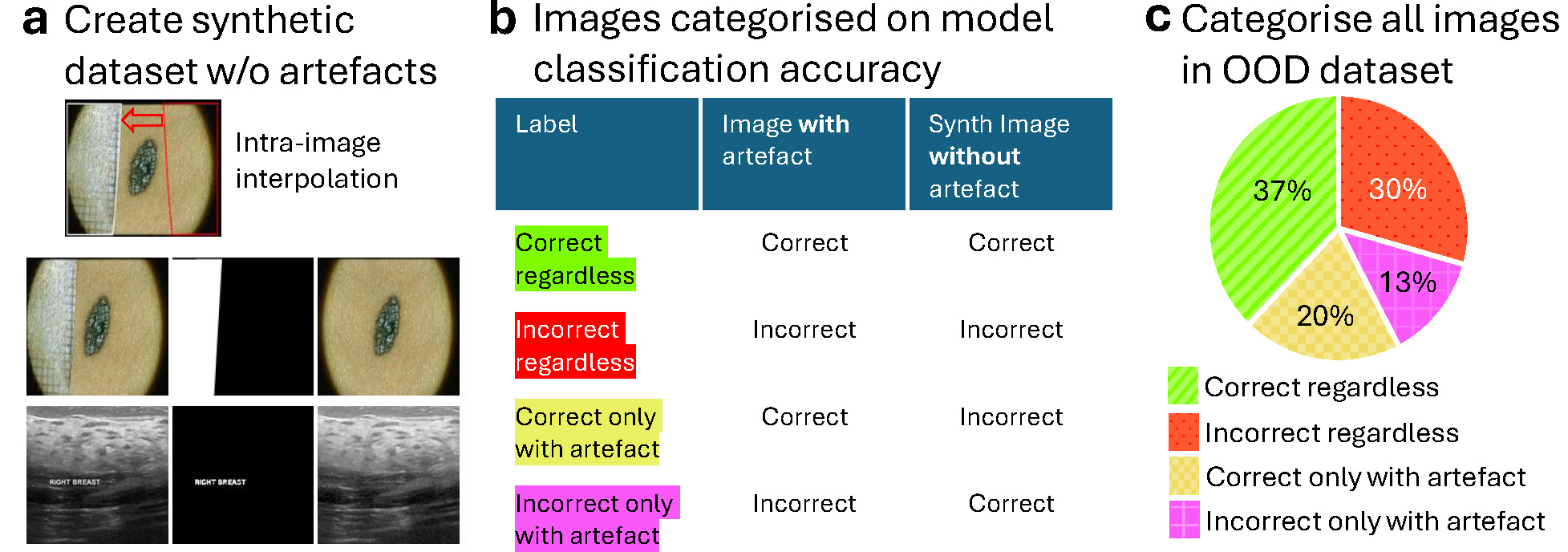} 
\caption{Workflow for categorising data based on the model prediction impact, using intra-image interpolation to create synthetic images without artefacts.} 
\label{Fig:workflow}
\end{figure}

Once the synthetic datasets were created, we compared the model's classification accuracy with the artefact (original image) and without the artefact. We categorised the OOD data depending if the diagnosis changed with the removal of the artefact (Fig. \ref{Fig:workflow}). After categorising the data, the next step involved integrating OOD detection methods into the DNN pipeline. The main focus was to evaluate which of these methods could effectively identify and dismiss potentially misleading predictions influenced by the OOD artefacts. To do this, we applied an OOD detection method and calculated the scoring function for each image in the ID and OOD test sets. We then set a threshold at the 75 percentile of the scoring function's for the held-out ID data $\lambda_{\text{ID-75}}$ \cite{hendrycks_baseline_2018}, removing all model predictions below this threshold. We did this for a confidence-based method (MCP) and feature-based method (Mahalanobis score). The purpose of this study was to determine which OOD methods made the predictions of the DNN more trustworthy, defined here as being more accurate (diagnoses are more often correct) and less likely to have its predictions influenced by OOD features (it can reliably dismiss or handle artefacts that it wasn’t trained with).

\section{Results}\label{Sec:Results}

We tested 16 OOD detection methods (described in Sec. \ref{Sec:Methods}) for the D7P and BreastMNIST OOD tasks, with results shown in Table \ref{tbl:results} - evaluating their performance for OOD detection (AUROC$_{\text{OOD}}$) and failure detection (AUROC$_f$). For MC dropout we used $p=0.3$ with 100 samples. Some methods like ODIN have hyperparameters, for which we show the optimised result as an upper bound. Previous works have shown that OOD artefacts, such as rulers, are optimally detectable in the early layers of a network, which are responsible for detecting low-level features \cite{anthony_use_2023}. So we applied Mahalanobis score and RMS on an early layer of the network (module 5), and we apply MBM on the first branch \cite{anthony_use_2023}.

\begin{table}[H]
\hspace{1.0em}
\caption{AUROC$_{\text{OOD}}$ and AUROC$_{f}$ (mean of 5 seeds) for OOD detection methods for a) D7P (ruler OOD) and b) BreastMNIST (annotation OOD) tasks. \textbf{Bold} highlights best result. * methods with hyperparameters optimised on OOD data.   }
\begin{tabular}{c |  >{\columncolor[gray]{0.93}}c |  >{\columncolor[gray]{0.93}}c | c | c |  >{\columncolor[gray]{0.93}}c |  >{\columncolor[gray]{0.93}}c | c | c }

\hline  \multirow{3}{*}{OOD-D method}
& \multicolumn{4}{c|}{D7P (ruler OOD)} & \multicolumn{4}{c}{BreastMNIST (anno. OOD)} \\
\cline{2-9} 
& \multicolumn{2}{c|}{\cellcolor[gray]{0.93}ResNet18 } & \multicolumn{2}{c|}{VGG16} & \multicolumn{2}{c|}{\cellcolor[gray]{0.93} ResNet18} & \multicolumn{2}{c}{VGG16} \\
\cline{2-9} 
& AUC$_{\text{OOD}}$ & AUC$_{f}$ & AUC$_{\text{OOD}}$ & AUC$_{f}$  & AUC$_{\text{OOD}}$ & AUC$_{f}$ & AUC$_{\text{OOD}}$ & AUC$_{f}$ \\
\hline 
 \multicolumn{9}{c}{\textit{Confidence-based Methods}} \\
\hline 
MCP \cite{hendrycks_baseline_2018} & 49.3  & 64.0 & 51.9 & \textbf{62.0} & 55.8 & 66.3 & 52.4 & 62.4 \\
SE \cite{hendrycks_scaling_2022} & 49.5 & 64.0 & 52.8 & \textbf{62.0} & 55.8 & 66.7 & 51.4 & 61.9 \\
MLS \cite{hendrycks_scaling_2022} & 48.6 & 63.7 & 51.5 & 61.9 & 57.9 & 69.7 & 52.4 & 64.2 \\
Energy Score \cite{liu_energy-based_2021} & 48.5 & 63.6 & 51.5 & 61.9 & 57.6 & 69.8 & 51.9 & 64.1 \\
MCDP-MCP \cite{gal_dropout_2016} & 49.3 & 64.0 & 52.0 & 61.8 & 55.8 & 66.3 & 51.9 & 62.2 \\
MCDP-PE \cite{jaeger_call_2022} & 49.5 & 64.0 & 51.7 & 61.9 & 55.8 & 66.7 & 50.3 & 62.2 \\
MCDP-MI\cite{jaeger_call_2022}  & 49.5 & 64.0 & 51.7 & 61.8 & 55.8 & 66.7 & 50.3 & 62.1 \\
DE-MCP \cite{lakshminarayanan_simple_2017}  & 49.9 & 64.2 & 52.7  & 61.9 & 56.0 & 66.4 & 53.3 & 62.3 \\
GradNorm \cite{huang_importance_2021} & 49.4 & 63.9 & 51.9 & 61.9 & 60.2 & 53.8 & 53.2 & 54.1 \\
ODIN* \cite{liang_enhancing_2020}  & 64.6 & 58.6 & 52.0 & \textbf{62.0} & 58.7 & 67.4 & 53.6 & 62.2 \\
ReAct* \cite{sun_react_nodate} & 67.2 & 60.6 & 61.5 & 58.9 & 60.2 & 65.2 & 58.0 & \textbf{64.4} \\
DICE* \cite{sun_dice_2022} & 68.5 & \textbf{67.8} & 57.7 & 59.2 & 58.0 & \textbf{70.9} & 59.1 & 64.0 \\
\hline
 \multicolumn{9}{c}{\textit{Feature-based Methods}} \\
\hline
Mahal. Score \cite{lee_simple_2018}  & 76.9 & 62.1 & 72.5 & 57.8 & 77.1 & 52.7 & 72.5 & 52.2 \\
MBM \cite{anthony_use_2023} & \textbf{80.7} & 61.7 & \textbf{73.8} & 56.8 & \textbf{77.4} & 53.9 & \textbf{76.8} & 52.0 \\
RMS  \cite{ren_simple_2021}  & 70.2 & 57.1 & 60.5 & 57.1 & 62.7 & 50.5 & 52.7 & 51.9 \\
GRAM \cite{sastry_detecting_2020} & 53.6 & 54.8 & 72.3 & 55.8 & 63.6 & 51.4 & 71.3 & 52.0 \\
\hline

\end{tabular}
\label{tbl:results}
\end{table}

\begin{figure}
\centering
 \includegraphics[width=9.25cm,height=5.25cm]{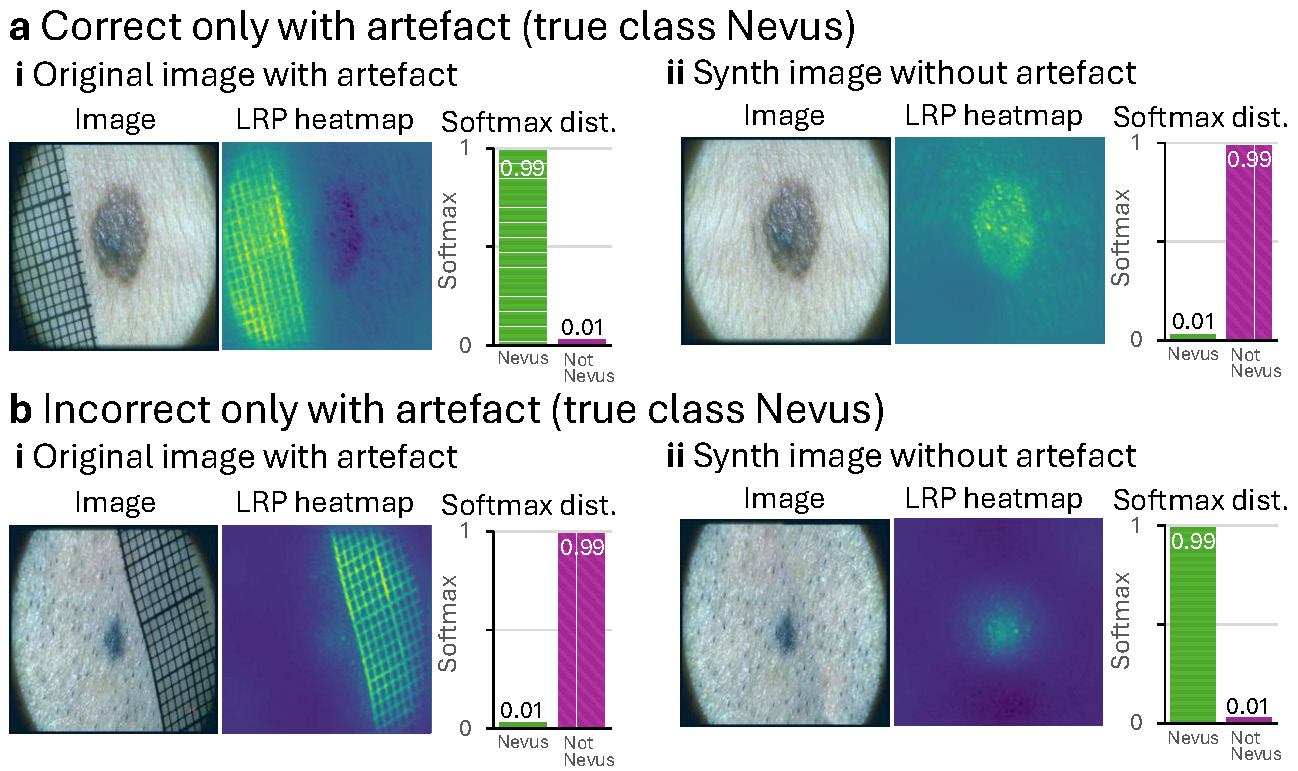}    
\caption{Comparison of the model's (VGG16) output and XAI heatmap (LRP) for D7P images with and without artefacts, showing cases where predictions are correct (a) or incorrect (b) only with the artefact. Used to demonstrate that OOD artefacts can lead to high confidence predictions.}
\label{Fig:clever_hans}
\end{figure}

From Table \ref{tbl:results}, it is observed that feature-based methods are typically more effective at detecting OOD inputs than confidence-based methods, quantified  using  AUROC$_{\text{OOD}}$. To explain why, we visualise the model predictions for two OOD images (both with and without the artefact) along with their eXplainable AI heatmaps using LRP (Fig \ref{Fig:clever_hans}).  We use this to challenge the assumption that  OOD artefacts will always cause a model to output a more uncertain (high entropy) output. The analysis shows that OOD artefacts can actually lead to high confidence predictions (high logit and hence softmax values), at comparable confidence to ID data. This phenomenon undermines the utility of confidence-based methods, that rely on the model's output layer, for detecting OOD inputs. There are several potential reasons for this phenomenon. One reason is the model can learn to identify correlations in the training data, such as specific intensity patterns in medical images. An OOD artefact that resembles these patterns can lead the model to make high-confidence predictions, even though the artefact is unrelated to the condition being diagnosed. Another cause is it has been theoretically demonstrated that ReLU networks inherently assign high confidence to images that are far from the training data \cite{Hein}. The results also show that feature-based methods perform comparatively worse at failure detection. This implies that it may be a false assumption to consider the Mahalanobis distance of an input to the training data as a reliable predictor of diagnostic accuracy \cite{lee_simple_2018} - we experimentally observe this phenomenon regardless of the network layer that the method is applied on (results not shown due to space constraints).

\begin{figure}[H]
\centering
 \includegraphics[width=10.5cm,height=8.95cm]{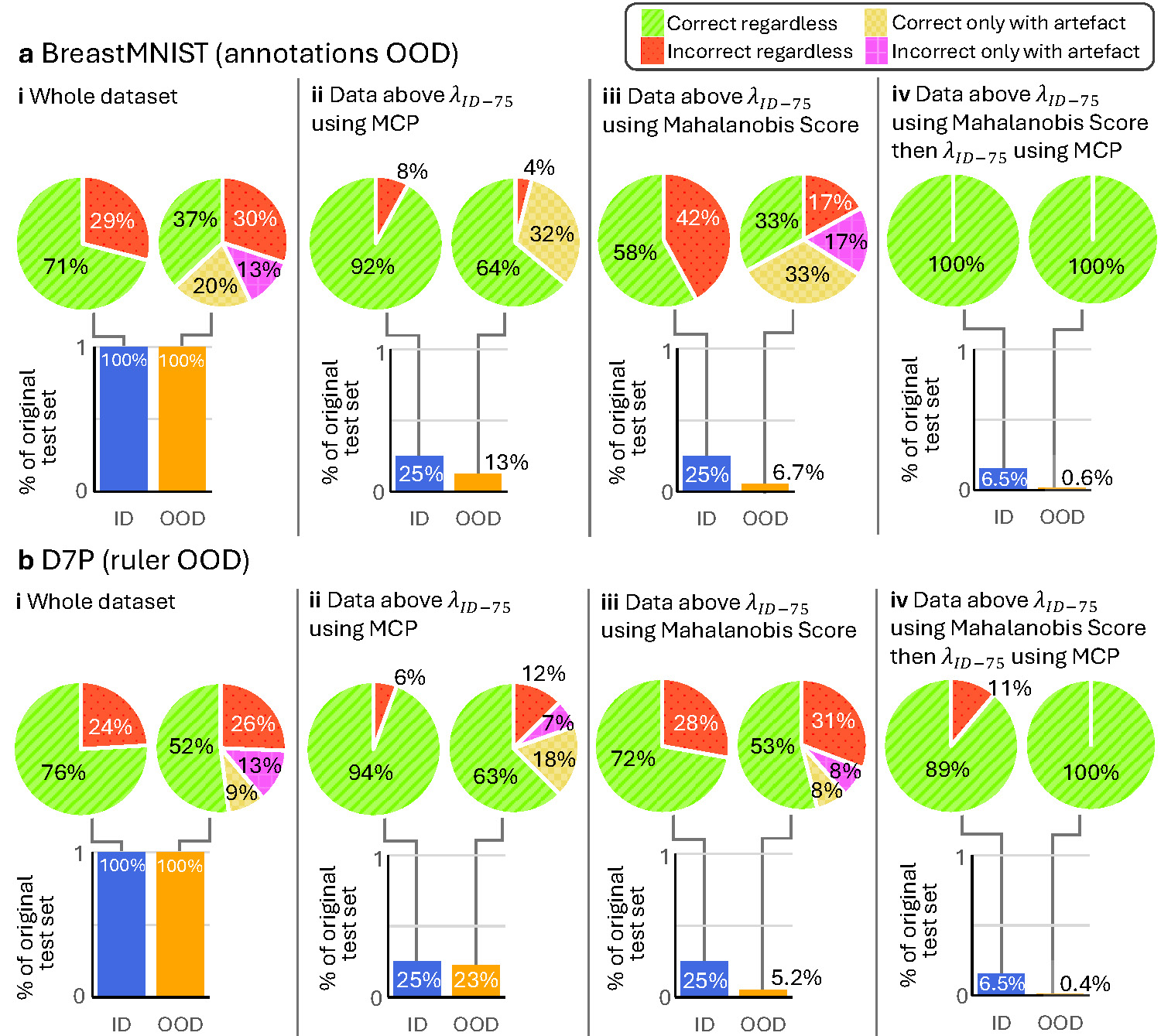}    
\caption{ a) BreastMNIST and b) D7P test sets were analysed using different OOD methods, removing predictions from a VGG16 model below $\lambda_{75-ID}$. The pie charts illustrate the distribution of predictions (see Fig. \ref{Fig:workflow}), while the bar charts display the percentage of ID and OOD data remaining after removing predictions below $\lambda_{75-ID}$, compared to the original dataset (i). The figure shows MCP's limitation in removing OOD data (ii) and Mahalanobis score's tendency to reduce prediction accuracy (iii). Combining these methods (iv) yields the most trustworthy predictions, but with a higher dismissal rate.
}

\label{Fig:threshold}
\end{figure}

Following that, we integrated a confidence-based method (MCP) and a feature-based method (Mahalanobis score) into a DNN pipeline, placing threshold $\lambda_{ID-75}$ and studying the predictions above the threshold (Fig. \ref{Fig:threshold}). 
We see that although the confidence-based method improves accuracy of predictions, it does not notably reduce the proportion of OOD images relative to ID images compared with the original dataset. For predictions on OOD images above $\lambda_{ID-75}$, the percentage which are correct only due to the presence of artefacts increases compared with the original dataset, which could cause an inflated AUROC$_f$ metric. This could give a misleading impression that applying MCP would lead to much more trustworthy predictions, when in fact these predictions are heavily impacted by the OOD artefact. This could be an issue if this results in an overconfidence in the model's ability to handle OOD data, masking its vulnerability and potentially leading to performance breakdowns post-deployment if the correlation between OOD artefacts and correct diagnoses changes. 
This also raises concerns if the failure detection framework for evaluating OOD methods is too simplistic, as these metrics don't consider cases where the model is correct for the wrong reasons.
We also see the feature-based method does better at reducing the number of OOD images, but the predictions above $\lambda_{ID-75}$ can have worse diagnosis accuracy compared with the original dataset (for both ID and OOD sets). Neither of these outcomes are ideal in terms of improving the trustworthiness of DNN predictions. If we consider feature-based methods as those capable of identifying images that look visually distinct from the training data (regardless of diagnosis accuracy) and confidence-based methods as those that are better at dismissing images with incorrect diagnoses (but struggle to identify visually distinct images), it motivates integrating both confidence and feature-based methods to compensate for their respective weaknesses. We test this by first removing predictions below $\lambda_{ID-75}$ for Mahalanobis score, then removing predictions below $\lambda_{ID-75}$ for MCP, to demonstrate that the remaining predictions are more accurate while reducing the number of predictions influenced by OOD features. Although it results in more predictions being dismissed, we argue that this configuration leads to more reliable DNN predictions. Hence, we suggest the community should consider systems which incorporate both a confidence-based and a feature-based method.

\section{Conclusion}
This paper sheds light into the weaknesses of current OOD detection methods. We show that confidence-based methods are less effective than feature-based methods in OOD detection, partly due to the false assumption that OOD artefacts consistently lead to higher model uncertainty.  This paper also explains that feature-based methods, while superior at identifying OOD inputs, under-perform in failure detection compared to confidence-based methods, which can reduce accuracy of predictions when integrated into a DNN pipeline. The paper suggests a step forward could be to seek combinations of confidence- and feature-based methods that compensate for their respective shortcomings.

\subsubsection{Acknowledgments.} HA is supported by a scholarship via the EPSRC Doctoral Training Partnerships programme [EP/W524311/1, EP/T517811/1]. The authors also acknowledge the use of the University of Oxford Advanced Research Computing (ARC) facility for the work (http://dx.doi.org/10.5281/zenodo.22558).

%
%
\bibliographystyle{splncs04_with_et_al}
\bibliography{references.bib}

\end{document}